\documentclass[10pt,twocolumn,letterpaper]{article}

\usepackage{iccv}
\usepackage{times}
\usepackage{epsfig}
\usepackage{graphicx}
\usepackage{amsmath}
\usepackage{amssymb}
\usepackage{cite}
\usepackage{booktabs}
\usepackage{pifont}
\usepackage{diagbox}
\usepackage{multirow}
\usepackage{graphicx}
\usepackage{array}
\usepackage{changepage}
\newcommand*{\affaddr}[1]{#1} 
\newcommand*{\affmark}[1][*]{\textsuperscript{#1}}
\newcommand*{\email}[1]{\texttt{#1}}

\usepackage[breaklinks=true,bookmarks=false]{hyperref}

\iccvfinalcopy 

\def\iccvPaperID{216} 

\begin{document}

\title{Neural Person Search Machines }

\author{%
	 Hao Liu \affmark[1]  \quad Jiashi Feng\affmark[2] \quad Zequn Jie\affmark[3]  \quad Karlekar Jayashree\affmark[4] \\ Bo Zhao \affmark[5] \quad Meibin Qi\affmark[1] \quad Jianguo Jiang\affmark[1] \quad Shuicheng Yan\affmark[6,2]\\	
	\affaddr{\normalsize \affmark[1]Hefei University of Technology} \quad
	\affaddr{\affmark[2]National University of Singapore} \quad
	\affaddr{\affmark[3]Tencent AI Lab} \quad\\
	\affaddr{\normalsize\affmark[4] Panasonic R\&D Center Singapore} \quad	
	\affaddr{\affmark[5]Southwest Jiaotong University} \quad	
	\affaddr{\affmark[6]360 AI Institute}\\
	\email{\small \{hfut.haoliu, zequn.nus\}@gmail.com, elefjia@nus.edu.sg, karlekar.jayashree@sg.panasonic.com}\\ \email{\small zhaobo@my.swjtu.edu.cn, \{qimeibin, jgjiang\}@hfut.edu.cn, yanshuicheng@360.cn}
}
\maketitle

\begin{abstract}
	We investigate the problem of person search in the wild in this work. Instead of comparing the query against all candidate regions generated in a query-blind manner, we propose to recursively shrink the search area from the whole image till achieving precise localization of the target person, by fully exploiting information from the query and contextual cues in every recursive search step. We develop the Neural Person Search Machines (NPSM) to implement such recursive localization for person search. Benefiting from its neural search mechanism, NPSM is able to selectively shrink its focus from a loose region to a tighter one containing the target automatically. In this process, NPSM employs  an internal primitive memory component to memorize the query representation which modulates the attention and augments its robustness to other distracting regions. Evaluations on two benchmark datasets, CUHK-SYSU Person Search dataset and PRW dataset, have demonstrated that our method can outperform current state-of-the-arts  in both mAP and top-1 evaluation protocols.     
\end{abstract}
\vspace{-1mm}
\section{Introduction}

Person search~\cite{zheng2016person,xiao2016end} aims to localize a specific  person matching the provided query in gallery images or video frames. It is a new and challenging task that requires  to address 
person detection and  re-identification simultaneously. 
It has many important applications in video surveillance and security, such as cross-camera visual tracking\cite{nino2016scalable}  and person verification\cite{Yang2016Large}.
But it is difficult in real-world scenarios due to various distracting factors including large appearance variance across multiple cameras, low resolution, cluttered background, unfavorable camera setting, \textit{etc}.  

\begin{figure}[t]
	\begin{center}
		\begin{tabular}{ccc}
			{\hspace{-3pt}}
			
			\includegraphics[width=0.95\linewidth]{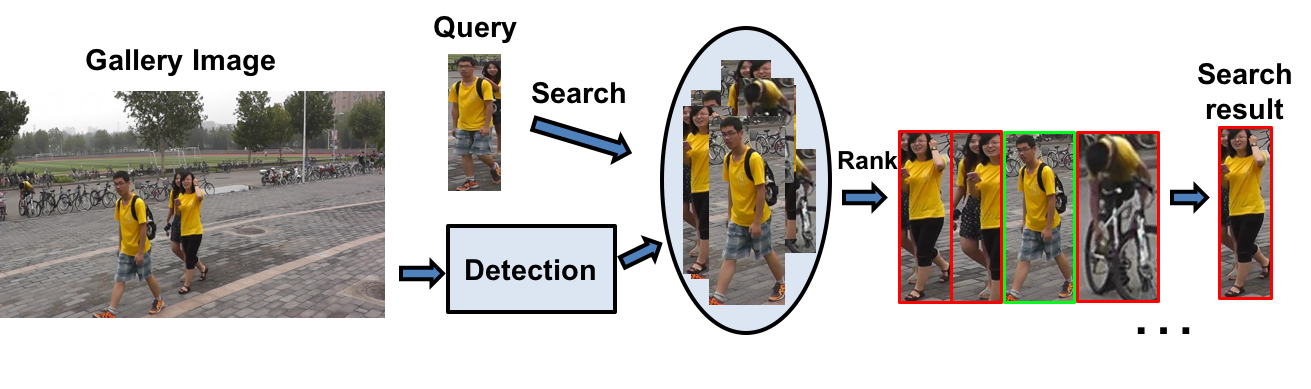}
			\vspace{-2.5mm}
			\\
			
			{\footnotesize{(a) Search process of previous methods}}
			\\
			\includegraphics[width=0.9\linewidth]{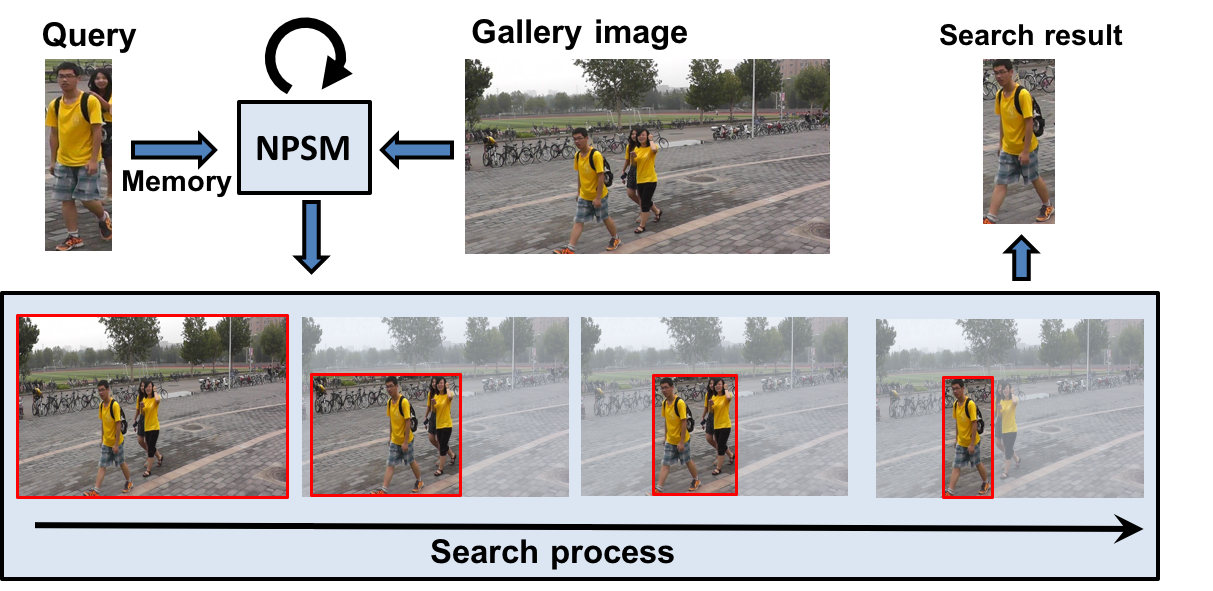}
			\\
			{\footnotesize{(b) Search process of the NPSM}}
			\vspace{-2.5mm}
		\end{tabular}
	\end{center}
	\caption{Demonstration of person search process for one gallery image in previous methods and our proposed method.  (a) The search process of  previous methods. The query person is one-by-one compared with the detection results for one gallery image; then the search result ranked at the first place is obtained. The red boxes indicate the wrong matched results while the green box represents the truly matched person. (b) The search process of the NPSM.  When a target person is searched within a whole scene, the search scope on which attention is focused is recursively shrinking with guidance from memory of the query person's appearance, which is marked in red boxes.}
	\label{fig:motivation}
	\vspace{-5mm}
\end{figure}

To date, only a few methods have been proposed  to address this task. In the pioneer work~\cite{xiao2016end}, Xiao \emph{et al.} adopted the end-to-end person search model based on the proposed Online Instance Matching (OIM) loss function to jointly train person detection and re-identification networks. The recent work~\cite{zheng2016person} also follows a similar pipeline. Generally, all of the previous person search methods are based on such a simple two-stage search strategy: first to detect all candidate persons within an image and then to perform exhaustive comparison between all possible pairs of the query and the candidates to output a search result ranked at the first place within the searched images. This pipeline has some drawbacks. Firstly, if the target person has distracting factors around, \emph{e.g.}, another person with similar appearance, the search accuracy would be affected by the distracting factors. Secondly, extra error, such as inaccurate detection, would be introduced by the two isolated frameworks, \textit{i.e.} person detection and re-identification.  See Figure~\ref{fig:motivation} (a) for demonstration. The red boxes indicate the wrong matched results while the green box represents the truly matched person.

For person search, it is commonly assumed that within an image, the target person only appears at a single location. Such instance-level exclusive cues imply that instead of examining all possible persons, a more effective strategy is to only search within the regions possibly \emph{containing the target person} in a coarse-to-fine manner. This is similar to human neural system for processing complex visual information\cite{anderson1990cognitive, Olshausen1993A}.  
More concretely, after seeing and remembering the appearance of a target person, one usually shrinks  his search area from a large scope to a small one  and performs matching with his  memory in details within the small scope with more effort. Such a coarse-to-fine search process is intuitively useful for existing person search solutions.

Inspired by above observations,  we propose a new and more effective person search strategy and develop the Neural Person Search Machines (NPSM). Compared to the search process in previous methods, our NPSM  (Figure~\ref{fig:motivation} (b)) takes the query person as memory to recursively guide the model to shrink the search region and judge whether the current region contains the target person or not.  This process would include more contextual cues beneficial for person matching. In Figure~\ref{fig:motivation} (b), the red box in each image from left to right corresponds to a region that can be focused on, and the arrow indicates a search process which can be considered as the continuous shrinkage of the focus region.  Additionally, those irrelevant regions can be ignored after every shrinkage of a subregion from a big region, which can reduce the interference of other unimportant regions. 

To model the above  person search process,  we need to solve  the following two non-trivial problems: 1) integrating information of the query person into the search process  as memory to exclude interference from impossible candidates; 2) judging which subregion should be focused on in the bigger region at each recursive step in the  coarse-to-fine search process under the guidance of memory.

For localizing the target person in a sequence correctly and fully exploiting the context information, we propose a neural search architecture to selectively concentrate on an effective subregion of the input region, and meanwhile ignore other perceived information from distracting subregions in a recursive way. Take the third subregion of the search process in Figure~\ref{fig:motivation} (b) for example, the proposed NPSM would highlight the truly matched person at the left side of the region and ignore the similar person at the right side. Considering the specific ability of Long Short-Term Memory (LSTM)~\cite{hochreiter1997long} to partially allow or deny information to flow into or out of its memory component, we build our Neural Search Networks (NSN) upon Convolutional LSTM (Conv-LSTM)~\cite{xingjian2015convolutional} units which are capable of preserving spatial information from the spatio-temporal sequences. 

Different from the vanilla Conv-LSTM, we augment our NSN by equipping it with external primitive memory that contains appearance information of the query and helps identify the candidate regions at the coarse level and discards irrelevant regions. The external primitive memory thus enables the query to be involved in the representation learning for person search as well as the recursive search process with region shrinkage. 

To sum up, in this work we go beyond the standard LSTM based models and propose a new person search approach called Neural Person Search Machines (NPSM) based on the Conv-LSTM~\cite{xingjian2015convolutional}, which contains the context information of each person and employs the external memory about the query person to guide the model to attend to the right region. Our approach is able to achieve better performance compared with other methods, as validated by experimental results. 

We make the following contributions to person search:
\begin{adjustwidth}{0mm}{0mm}
	\noindent1) We redefine the person search process as a detection free procedure of recursively focusing on the right regions.
	
	\noindent2) We coin a novel method  more robust to distracting factors benefiting from contextual information.
	
	\noindent3) We propose a new neural search model that can integrate the query person information into primitive memory to guide the model to recursively focus on the effective regions.
\end{adjustwidth}

\section{Related Work}
Person search can be regarded as the combination of person re-identification and person detection. 
Most of existing works of person re-identification focus on  designing hand-crafted discriminative features~\cite{farenzena2010person,gray2008viewpoint,Liao2015Person},  learning deep learning based high-level features~\cite{li2014deepreid, xiao2016learning, ahmed2015improved, wu2016personnet, liu2016end, liu2017video} and learning distance metrics~\cite{tao2016person, zhang2016learning, liu2015kernelized,K2012large,li2015multi}. 
Recent deep learning based person re-identification methods\cite{li2014deepreid,ahmed2015improved,liu2016end, liu2017video} 
re-design the structure of the deep network to improve performance. For instance, \cite{ahmed2015improved} designed two novel layers to capture relationships between two views of a person pair. Among distance metric learning methods, \cite{K2012large} proposed KISSME (KISS MEtric) to learn a distance metric from equivalence constraints. Additionally, \cite{zhang2016learning} proposed to solve the person re-id problem by learning a discriminative null space of the training samples while \cite{li2015multi} proposed a method learning a shared subspace across different scales to address the low resolution person re-identification problem.

For person detection, Deformable Part Model (DPM)~\cite{felzenszwalb2010object}, Aggregated Channel Features (ACF)~\cite{dollar2014fast} and  Locally Decorrelated Channel Features (LDCF)~\cite{Nam2014Local} are three representative methods relying on hand-crafted features and linear classifiers to detect pedestrians. Recently, several deep learning-based frameworks have been proposed. In \cite{tian2015deep},  DeepParts was proposed to handle occlusion with an extensive part pool. Besides, \cite{cai2015learning} proposed the CompACT boosting algorithm learning complexity-aware detector cascades for person detection. In our knowledge, two previous works~\cite{xiao2016end,zheng2016person} address person search by fusing person re-identification and detection into an integral pipeline to consider whether any complementarity exists between the two tasks. \cite{xiao2016end} developed an end-to-end person search framework to jointly handle both aspects with the help of Online Instance Matching (OIM) loss while \cite{zheng2016person} proposed ID-discriminative Embedding (IDE) and Confidence Weighted Similarity (CWS) to improve the person search performance. However, these two works simply focus on how the interplay of pedestrian detection and person re-identification affects the overall performance, and they still isolate the person search into two individual components (detection and re-identification),  which would introduce extra error, \emph{e.g.} inaccurate detection.  In this paper, we regard person search as a detection-free process of gradually removing interference or irrelevant target persons for the query person.

Recently, LSTM based attention methods have shown good performance in image description~\cite{xu2015show}, action recognition~\cite{li2016videolstm,sharma2015action}
and person re-identification~\cite{liu2016end}. In \cite{xu2015show}, Xu~\textit{et al.}\ showed how
the learned attention can be exploited to give more interpretability into the model generation process, while \cite{sharma2015action, li2016videolstm} adopted attention methods to recognize important elements in video frames based on the action that is being performed. Moreover, \cite{liu2016end} formulated an attention model as a triplet recurrent neural network which dynamically generates comparative attention location maps for person re-identification. Analogously, our proposed NPSM also has such a locally emphasizing property, but NPSM is a query-aware model while the above attention-based methods all adopt a blind attention mechanism. 


\vspace{-1.5mm}
\section{Proposed Neural Person Search Machines}
In this section, we present the architecture details of the proposed Neural Person Search Machines (NPSM), and explain how it works with the primitive memory modeling to facilitate person search.
\vspace{-0.9mm}
\subsection{Architecture Overview  }
\begin{figure*}[htb]
	\centering
	\includegraphics[width=14.5cm,height=7.9cm]{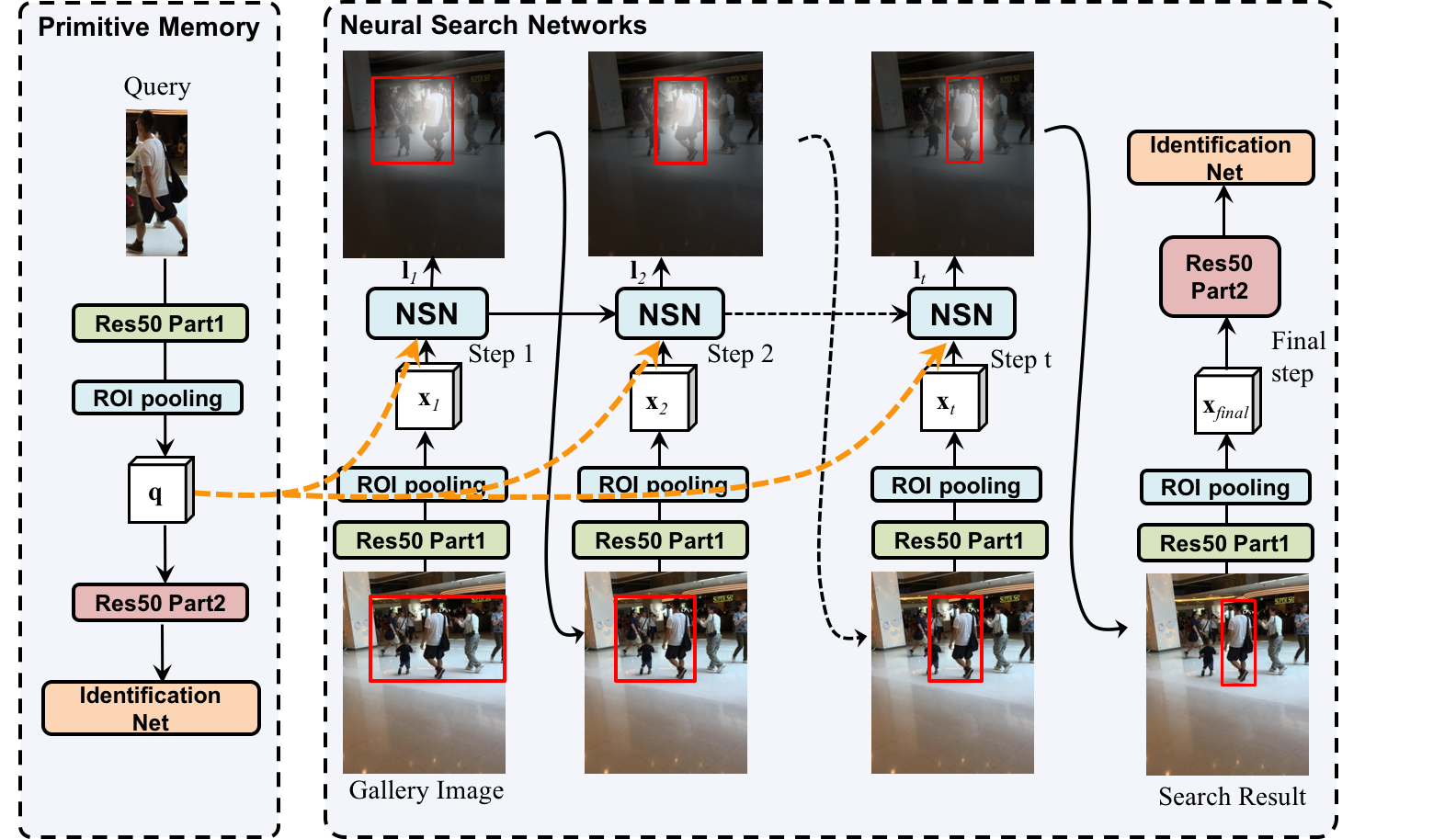}
	\caption{Architecture of our proposed Neural Person Search Machines (NPSM). It consists of two components, \textit{i.e.} Primitive Memory and Neural Search Networks. It works by recursively shrinking the search area from the whole image to the precise bounding box of the person of interest under the guidance (orange dotted lines) of Primitive Memory. And each region in the shrinking search process would contain the contextual information of the final search result. Red boxes denote the shrinking regions highlighted at different recursive steps in our NPSM. ``Res50 Part1'' corresponds to the \textit{conv1} to \textit{conv4\_3} of ResNet-50 while ``Res50 Part2'' represents the \textit{conv4\_4} to \textit{conv5\_3} of ResNet-50. Best viewed in color.}
	
	\vspace{-3.5mm}	
	\label{fig:overview}
\end{figure*}

\vspace{-1mm}
The overall architecture is shown in  Figure~\ref{fig:overview}. It consists of two components, \textit{i.e.} Primitive Memory and Neural Search Networks. We propose to solve person search  by recursively shrinking the search area from the whole image to the precise bounding box of the person of interest. And each region in the shrinking search process would contain the contextual information of the final search result. Besides recursively utilizing the contextual cues, NPSM provides extra robustness to interference from other distracting subregions for the model in the search process.

The proposed NPSM is trained end-to-end to learn to make a decision on the subregion attention at each recursive step and finally localize the person of interest. The Neural Search Network enables the model to automatically focus on relevant regions, and the Primitive Memory that models the representation of the query person continuously provides extra cues for every search step and facilitates more precise localization of persons.  After performing the recursive region shrinkage, the model reaches a search result with the biggest confidence as the final search result with an gallery image. 
Note that, different from previous works~\cite{xiao2016end,zheng2016person}, our method is detection-free and includes no Non-Maximum Suppression (NMS), as it is a search process performing simultaneous region shrinking and person identification. When the searching is finished, there will be only one bounding box person search result left. 

\subsection{Person Search with NPSM}

As aforementioned, we redefine the person search process as the recursive region shrinking process. It is equivalent to recursively focusing on a subregion containing the person of interest from a bigger region. Here we describe the details of our proposed NPSM and explain how to perform the recursive region shrinking to search the target person for each gallery image.

\vspace{-2.5mm}
\subsubsection{Neural Search Networks}
\vspace{-1mm} 
Learning to search for a person from a big region to a specific person region within the gallery image can be deemed as a sequence modeling problem. Specifically, the shrinking regions constitute a sequence. Thus a natural choice for the model candidates is the Recurrent Neural Network (RNN) or LSTM based RNN. However, the vanilla LSTM\cite{hochreiter1997long} only models sequence information through fully connected layers and requires vectorizing 2D feature maps.  This would result in the loss of spatial information, harming  person localization performance. In order to preserve the spatial structure of the regions over the shrinking process shown in Figure~\ref{fig:overview}, we design a new network called Neural Search Network (NSN) based on Convolutional LSTM (Conv-LSTM)~\cite{xingjian2015convolutional} for each recursive step.  Conv-LSTM replaces the fully connected multiplicative operations in an LSTM unit with convolutional operations. Different from it, the NSN has an additional memory component recording the query.  


Conv-LSTM can be used for building attention networks that can learn to pay attention to critical regions within feature maps. Thus, Conv-LSTM based NSN is also equipped with attention mechanism to learn to gradually shrink the region and selectively memorize the contextual information contained in the searched bigger region at each recursive step. However, our neural search model has a unique feature that distinguishes it from a plain attention model:~in addition to gallery images, a query illustrating the search target is also input to the search network. Traditional attention networks cannot well model such extra cues. In this work, we propose to model such query information into the primitive memory in order to facilitate person search.  

We now  elaborate on the new Neural Search Networks (NSN) of our NPSM, tailored for the person search task. In the NSN component, the query person information, denoted as $\mathbf{q}$, is integrated into the computation within gates and cell states in a way to bias the updating of internal states towards emphasizing information relevant to the query while forgetting irrelevant information.  Here the query feauture $\mathbf{q}$ is extracted from the query image through the pre-trained ``Res50 part1'' (\textit{conv1} to \textit{conv4\_3} of ResNet-50\cite{he2016deep}) which is the same as the one extracting features from gallery images. The cell and gates in the NSN are defined as
\vspace{-2.5mm}
\begin{align}
	\small
	\mathbf{i}_{t} &=\sigma \left( \mathbf{w}_{xi} \ast \mathbf{x}_t + \mathbf{w}_{hi} \ast \mathbf{h}_{t-1} +\mathbf{w}_{qi} \ast \mathbf{q}+ b_i\right) \nonumber\\  
	\mathbf{f}_{t} &=\sigma \left( \mathbf{w}_{xf} \ast \mathbf{x}_t + \mathbf{w}_{hf} \ast \mathbf{h}_{t-1} +\mathbf{w}_{qf} \ast \mathbf{q}+ b_f\right) \nonumber \\ 
	\mathbf{o}_{t} &=\sigma \left( \mathbf{w}_{xo} \ast \mathbf{x}_t + \mathbf{w}_{ho} \ast \mathbf{h}_{t-1}+\mathbf{w}_{qo} \ast \mathbf{q} + b_o\right) \label{NSN}\\ 
	\mathbf{g}_{t} &=\textrm{tanh}\left(\mathbf{w}_{xc} \ast \mathbf{x}_t + \mathbf{w}_{hc} \ast \mathbf{h}_{t-1} +\mathbf{w}_{qc} \ast \mathbf{q}+ b_c\right) \nonumber\\
	\mathbf{c}_{t} &= \mathbf{f}_{t} \odot \mathbf{c}_{t-1} + \mathbf{i}_{t} \odot \mathbf{g}_{t} \nonumber \\
	\mathbf{h}_t &= \mathbf{o}_{t} \odot \textrm{tanh}\left(\mathbf{c}_{t}\right), \nonumber
\end{align}
\vspace{-5.5mm}
\\
where $\ast$ represents the convolutional operation and $\odot$ is the Hadamard product, $\mathbf{w}_{x\sim}$, $\mathbf{w}_{h\sim}$ are two-dimensional convolutional kernels and $\mathbf{x}_t$  which is the feature map of the region highlighted by the previous time-step denotes the input at time step $t$.  The input gate, forget gate, output gate,  hidden state and memory cell are denoted as  $\mathbf{i}_t$, $\mathbf{f}_t$, $\mathbf{o}_t$, $\mathbf{h}_t$, $\mathbf{c}_t$  respectively, which are all three-dimensional tensors retaining spatial dimensions. With their control, the contextual information can be selectively memorized. Note the query person information $\mathbf{q}$ is independent of the time step $t$, therefore serving as the global primitive memory that  guides  the person search procedure continuously. The effect of such memory information over the states is modeled through the parameter $\mathbf{w}_{q \sim}$. 
\vspace{-1.5mm}
\subsubsection{ Region Shrinkage with Primitive Memory}
\vspace{-1mm}
As stated above, the goal of NPSM is to effectively shrink regions containing the target person based on the neural search mechanism, guided by the primitive memory. That is, the NPSM will decide which local region should be focused on at each recursive step in the search process as shown in Figure~\ref{fig:overview}. Through this way, more context information would be included from a large region and the number of irrelevant person candidates with the target person would be recursively reduced in the search process. In this subsection, we introduce how the subregion of each recursive time-step is generated and shrunk from the bigger region of the previous time-step.

Here we define the region covered by the highlighted proposal bounding boxes induced by current attention map as follows:
\vspace{-0.3mm}
\begin{equation*}
	\small
	\textbf{R}= (\min(\boldsymbol{\theta}_{x1}),\min(\boldsymbol{\theta}_{y1}),
	\max(\boldsymbol{\theta}_{x2}),\max(\boldsymbol{\theta}_{y2})),
	\label{bounding}
\end{equation*} 
where $\boldsymbol{\theta}_{x1}, \boldsymbol{\theta}_{y1},\boldsymbol{\theta}_{x2},\boldsymbol{\theta}_{y2}$ are the top left and lower right corner coordinates of all the highlighted bounding boxes from a predefined collection, generated by an unsupervised object proposal model (\textit{e.g.}, Edgeboxes\cite{zitnick2014edge}). 
Then we separate the region \textbf{R} into several candidate subregions according to the relationship of each contained bounding box in the region \textbf{R}. In this paper, we choose the Euclidean distance as the metric of the relationship defined as 
\vspace{-1.5mm} 
\begin{align}
	\small
	d(\mathbf{a},\mathbf{b})&=\sqrt{\sum\nolimits _{i=1}^{2}(a_i-b_i)^2},\label{{metric}}
\end{align}	
 where	$\mathbf{a}$ and $\mathbf{b}$ are the centre coordinates of two proposal bounding boxes \textit{A} and \textit{B} respectively. Specifically, $\mathbf{a}=\left( a_1, a_2\right)$, $\mathbf{b}=\left( b_1, b_2\right)$, $a_1= a_{x1}+0.5\left( {a_{x2}} -  {a_{x1}}\right)$, $a_2= a_{y1}+0.5\left( {a_{y2}} - {a_{y1}}\right)$, $b_1= b_{x1}+0.5(b_{x2} - b_{x1}), b_2= b_{y1}+0.5(b_{y2} - b_{y1})$. $(a_{x1}, a_{y1})$ and $(a_{x2}, a_{y2})$ are the top left and lower right coordinates of bounding box \textit{A} while $(b_{x1}, b_{y1})$ and $(b_{x2}, b_{y2})$ are the top left and lower right coordinates of bounding box \textit{B}.  Then the proposal bounding boxes can be grouped into $C$ clusters according to their relationships.  The corresponding  subregions covered by proposals are $\textbf{R}_{(C)}^{sub}$. We denote the parent region to generate subregions $\textbf{R}_{(C)}^{sub}$ as $\textbf{R}^{par}$.

At each recursive step $t$, the proposed NSN outputs an attention map which predicts the scores (reflecting confidence on containing the target person given the primitive memory information) of shrinking to region $\textbf{R}_{t, (C)}^{sub}$  after NSN taking input the parent region $\textbf{R}_{t-1}^{par}$ at the previous step $t-1$. 

More specifically, at each time step (corresponding to shrinking to one region), NSN takes input the query person feature \textbf{q} and the region feature $\mathbf{x}_t$ extracted from pre-trained ``Res50 part1'' which denotes the \textit{conv1} to \textit{conv4\_3} of ResNet-50\cite{he2016deep}. Here, we add a Region of Interest (ROI) pooling layer following ``Res50 part1'' to make sure  the regions of different sizes can have feature maps of the same size $K \times K \times D$.   
Compared with the standard LSTM based model relying on multi-layer perceptron, NSN uses convolutional layers to integrate the region representation with primitive memory and produce attention maps. Specifically, at each time step $t$, an attention score map of size $K \times K$ for $K \times K$ locations is computed:
\begin{align}
	\small
	\mathbf{z}_t &=\mathbf{w}_z \ast \textrm{tanh} \left( \mathbf{w}_{qa} \ast \mathbf{q} + \mathbf{w}_{ha} \ast \mathbf{h}_t + b_a \right) \label{eq_att1}\\
	\mathbf{l}_{t}^{i,j}  &= \frac{\exp(\mathbf{z}_{t}^{ij})}{\sum_{i}\sum_{j}\exp(\mathbf{z}_{t}^{ij})}.
	\label{eq_att2}
\end{align}
\vspace{-3mm}
\\The score for location $(i,j)$ is denoted as $\mathbf{l}_{t}^{i,j}$.

Then, in the process of region shrinkage, the NSN computes the average scores of different subregions from  the parent region. NSN highlights the subregion with the maximum score as the region to be searched in the next step. This computation would be performed many times until the search path reaches the final proposal. The average score of the subregion is computed as follows:
\vspace{-2.5mm}
\begin{equation}
	\small
	\mathbf{S}_t =\frac{1}{m \cdot n}\sum_{i=1}^{m}\sum_{j=1}^{n}\mathbf{l}_{t}^{i,j},
	\label{eq_score}
	\vspace{-1mm}
\end{equation} 
where $m$ and $n$ are the height and the width of the subregion respectively. $\mathbf{l}_{t}^{i,j}$ corresponds to the score map defined in Eqn.~\eqref{eq_att2}  generated on the parent region. In other words, our model does not stick to the single region. If some regions not highlighted before receive higher attention at certain search step, our model would jump to that region with higher intra-region confidence scores. In this way, the accumulative error in the shrinkage process can be alleviated. Note that our NSN serves as a region shrinkage method. In other words, our NSN only outputs the most similar proposal with the query person in each gallery image. Therefore, the features of the query person image and the final search result are extracted from the ``Identification Net'' (orange boxes in  Figure~\ref{fig:overview}) of the trained model when the searching is finished. Here, the ``Identification Net'' takes input the output of ``Res50 Part2'' (pink boxes in Figure~\ref{fig:overview}) representing the \textit{conv4\_4} to \textit{conv5\_3} of ResNet-50. And it consists of a global average pooling layer and a 256-dimension Fully Connected layer. Then the \textit{cosine} similarity between the features of the query person and the final person search result is computed for evaluation.

\vspace{-1.1mm}
\subsection{Training Strategy}
\vspace{-0.5mm}

Here we detail the training of the proposed model. Firstly, we use the architecture proposed in OIM\cite{xiao2016end} to \emph{pre-train} the  Fully Convolutional Networks (FCN) including both ``Res50 part1'' and ``Res50 part2'. Then for the region at each recursive time-step, the feature is extracted from the ROI pooling layer after the pre-trained ``Res50 part1''. After that, all the features are fed to the NSN and we add a  convolutional layer of size  $1 \times 1 \times 2$ after output of each time step to calculate the ``region shrinkage loss''. Here we adopt segmentation alike softmax loss as the ``region shrinkage loss''. The supervision label of each time step is defined as 
\begin{equation}
	\small
	\mathbf{U}_{t} = 
	\begin{cases}
		\textbf{1}, \text{if } G \in R_{t}  \\
		\textbf{0}, \text{otherwise}, \\ 
	\end{cases}
\end{equation}
\vspace{-3mm}
\\where $G$ is the ground truth bounding box of the target person while $R_{t}$ is the region box reached at the $t$th time step. This training strategy enables the proposed network to produce proper attention maps that fall into the region containing the target person as tight as possible. In other words, our NPSM is expected to predict the probability of the target person appearing at each location in a gallery image. 

Besides, to make the learned feature more discriminative, we add an identification loss following the ``Identification Net'', which takes input the output feature \textbf{u} of ``Identification Net'' and is defined as
\begin{align}
	\small
	P(z=c|\mathbf{u})&=\frac{\exp(S_{c}\mathbf{u})}{\sum_{k}\exp(S_{k} \mathbf{u})},\\
	L_{iden} &= -\textup{log}(P(z=c|\mathbf{u})).
\end{align}
where there are a total of \textit{I} identities, \textit{z} is the identity of the person, and \textit{S} is the softmax weight matrix while $S_{c}$ and $S_{k}$ represent the \textit{c}th and \textit{k}th column of it, respectively.

\section{Experiments}
\vspace{-1mm}
\subsection{Datasets and Evaluation Protocol}
\vspace{-1mm}
\subsubsection{Datasets}
\vspace{-1mm}
\textbf{CUHK-SYSU: }CUHK-SYSU~\cite{xiao2016end} is a large-scale person search dataset with diverse scenes,  containing 18,184 images, 8,432 different persons, and 96,143 annotated pedestrians bounding boxes. 
Each selected query person appears in at least two images
captured from different viewpoints. The images present large variations in viewpoint, lighting, resolution, occlusion and background, intensively reflecting the real application scenarios and scene diversity. We use the official training/test split provided by the dataset. The training set contains 11,206 images and 5,532 query persons. Within the testing set, the query set contains 2,900 persons and the gallery contains  6,978 images in total.  

\noindent\textbf{PRW: }The PRW dataset~\cite{zheng2016person} is extracted from one 10-hour video captured  on a university campus. The dataset includes 11,816 video frames of scenes captured by 6 cameras. 
In total 11,816 frames are manually annotated, giving  43,110 pedestrian
bounding boxes. Among them, 34,304 pedestrians are annotated with 932 IDs. It provides 5,134 frames of 482 different persons for training. The provided testing set
contains 2,057 query persons and a gallery  of 6,112 images.
\vspace{-7mm}
\subsubsection{Evaluation Protocol}
\vspace{-1mm}
We adopt  the mean Averaged Precision (mAP) and the top-1 matching rate as performance metrics, which are also used in OIM\cite{xiao2016end} and \cite{zheng2016person}. Using the mAP metric, person
search performance is evaluated in a similar way as  detection, reflecting the accuracy of detecting the query
person from  the gallery images. 
The top-1 matching rate treats person search as a ranking
and localization problem. A matching is counted if a bounding box among
the top-1 predicted boxes overlaps with the ground truth 
larger than the threshold 0.5.  

\subsection{Implementation Details}
\vspace{-1mm}
In this paper, the Fully Convolutional Networks (FCN) including both ``Res50 part1'' and ``Res50 part2'' are \emph{pre-trained} by using the architecture proposed in OIM\cite{xiao2016end}. For the input region at each time-step, we apply an ROI pooling layer on the $conv4\_3$ convolutional features of  it
to normalize all the features to the  same size of 14 $\times$ 14 $\times$ 1024. For query person images, we also  extract  their 14 $\times$ 14 $\times$ 1024 convolutional features in the same way. These features are then fed into the NPSM architecture. In particular,  within NSN,  the  convolutional kernels  for input-to-input states and state-to-state transitions are fixed as 3 $\times$ 3 with 1024 channels. At each recursive search step, we set the number $C$ of  subregions covered by clustered proposals to 3. We implement our network using Theano \cite{2016arXiv160502688short} and Caffe \cite{jia2014caffe} deep learning framework. The training of the NPSM converges in roughly 50 hours for CUHK-SYSU dataset and 40 hours for PRW dataset on on a machine with 64GB memory, NVIDIA GeForce GTX TITAN X GPU and Intel i7-4790K Processor. The initial learning rate is 0.001 and decays at the rate of 0.9 for the weight updates of RMSProp~\cite{tieleman2012lecture}. Additionally, we manually augment the data by performing random 2D translation. The speed of our method is close to real-time. For one gallery image, our model takes round 1s to output a final searched result. However, overhead of ranking over gallery is dominating. For the CUHK-SYSU with gallery size of 100, calculating cosine similarity between search result from all the gallery images and query for ranking takes round 20s. For the PRW with 6,112 gallery images, ranking over gallery takes round 15 minutes.

\subsection{Ablation Study}
In this subsection, we perform several analytic experiments on CUHK-SYSU benchmark to investigate the contribution of each component in our proposed NPSM architecture. We analyze attention prediction, contextual cue and primitive memory of query person. In total we have three variants by training the model based on different combinations of the above factors. And the gallery size is set to 100. The details and corresponding results are shown in Table~\ref{tab:alb_st}.

As aforementioned, we employ the framework in OIM~\cite{xiao2016end} which involves none of three factors, as the baseline. Based on this framework, the results of OIM\cite{xiao2016end} are obtained. In the method named ``NPSM w/o C'', we remove the ``contextual cue and primitive memory integration'' part (corresponding to Eqn.~\eqref{NSN}) of the NSN in our proposed NPSM. Instead, at each recursive step, we replace the ``contextual que and primitive memory integration'' part with a 3 $\times$ 3 $\times$ 1024 convolutional layer followed by the concatenation of the FCN (``Res50 part1'') feature map  of the query person (primitive memory) and the current step region (\textbf{q} and $\mathbf{x}_t$). Moreover, for each recursive step, we only keep the shrinking region generation method and the attention score prediction model (Eqn. \eqref{eq_att2} and \eqref{eq_score} ) to predict the attention score map. This setting makes our NPSM a simple version without contextual cues involved but still with the attention prediction ability.  Furthermore, in the method named ``NPSM w/o A\&C'', we further remove the attention prediction model and only generate the shrinking region as the input of each recursive step and add a 1024-dimension fully connected (FC) layer and a 2-dimension FC layer after the output (concatenation of the FCN feature map of query person (primitive memory) and the current region) of each recursive step. And the 2-dimension FC layer aims at predicting the score of each highlighted subregion from the parent region. From comparison between the results of ``OIM'' and ``NPSM w/o A\&C'', we can see that simply using primitive memory of query without contextual cues involved to search for a target person in the recursive way can not achieve satisfactory results. From the result of ``NPSM w/o C'' , we find that the sightly higher performance is achieved than the baseline due to usage of the attention model which can introduce more spatial location information than the ``NPSM w/o A\&C''. However, both ``NPSM w/o A\&C'' and ``NPSM w/o C'' lack a contextual cue memory mechanism. In other words, the above methods are unable to memorize the context information provided in a larger region through previous recursive steps. From the result of  ``NPSM''  overtaking the baseline method ``OIM'' by 2.4\% and 2.5\% for mAP and top-1 evaluation protocol,  we find that the neural search mechanism induced by our proposed NPSM is beneficial for person search performance, and memory of query person can also effectively guide the neural search model to find the right person.

\begin{table}[t]
	\centering
	\small		
	\newcommand{\tabincell}[2]{\begin{tabular}{@{}#1@{}}#2\end{tabular}}
	\caption{Results of ablation study on CUHK-SYSU dataset with 100 gallery size setting. Legend: \textbf{A}: Attention prediction model, 
		\textbf{C}: Contextual cue, \textbf{P}: Primitive memory of query person.  ``w/o A\&C' and ``w/o C'' are short for ``without Attention prediction model and Contextual cue'' and ``without Contextual cue but with Attention prediction model'' respectively.} 
	\label{tab:alb_st}%
	\begin{tabular}{@{}l|cccccc}
		\toprule
		& \textbf{A}     & \textbf{C}     & \textbf{P}         & \textbf{mAP(\%)}   & \textbf{top-1(\%)} \\
		\midrule
		OIM (Baseline)  &\ding{53}       &\ding{53}       &\ding{53}           &  75.5     & 78.7 \\
		\midrule
		NPSM w/o A\&C &\ding{53}       &\ding{53}       &\ding{51}               &56.5      & 62.5 \\
		NPSM w/o C&\ding{51}       &\ding{53}       &\ding{51}              &  72.5     & 76.3 \\
		
		\textbf{NPSM} &\ding{51}       &\ding{51}       &\ding{51}            & \textbf{77.9} & \textbf{81.2} \\
		\bottomrule
	\end{tabular}%
	\vspace{-3mm}
\end{table}%

\subsection{Comparison with State-of-the-art Methods}
We compare NPSM with several state-of-the-arts, including end-to-end person search ones proposed by Xiao \textit{et al.}~\cite{xiao2016end} and Zheng \textit{et al.}~\cite{zheng2016person} and some other methods combining commonly used pedestrian detectors (DPM\cite{felzenszwalb2010object}, ACF\cite{dollar2014fast}, CCF\cite{Yang2015Convolutional}, LDCF\cite{Nam2014Local} and their respective R-CNN\cite{girshick2016region}) with hand-crafted features (BoW\cite{Zheng2015Scalable}, LOMO\cite{Liao2015Person}, DenseSIFT-ColorHist (DSIFT)\cite{Zhao2013Unsupervised}) and distance metrics (KISSME\cite{K2012large}, XQDA\cite{Liao2015Person}).

\begin{figure}[htb]
	\centering
	\includegraphics[width=0.55\linewidth]{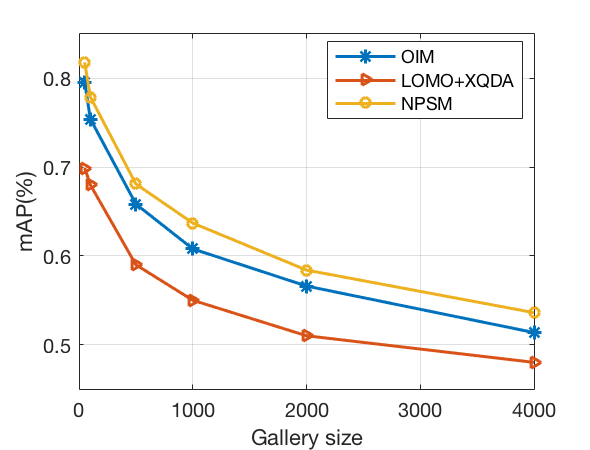}
	\vspace{-0.5mm}	
	\caption{Test mAPs of different approaches under different gallery sizes.} 
	
	\vspace{-2.6mm}	
	\label{fig:gallerysize}
\end{figure}

\begin{figure*}[htb]
	\centering
	\includegraphics[height=8cm,width=17cm]{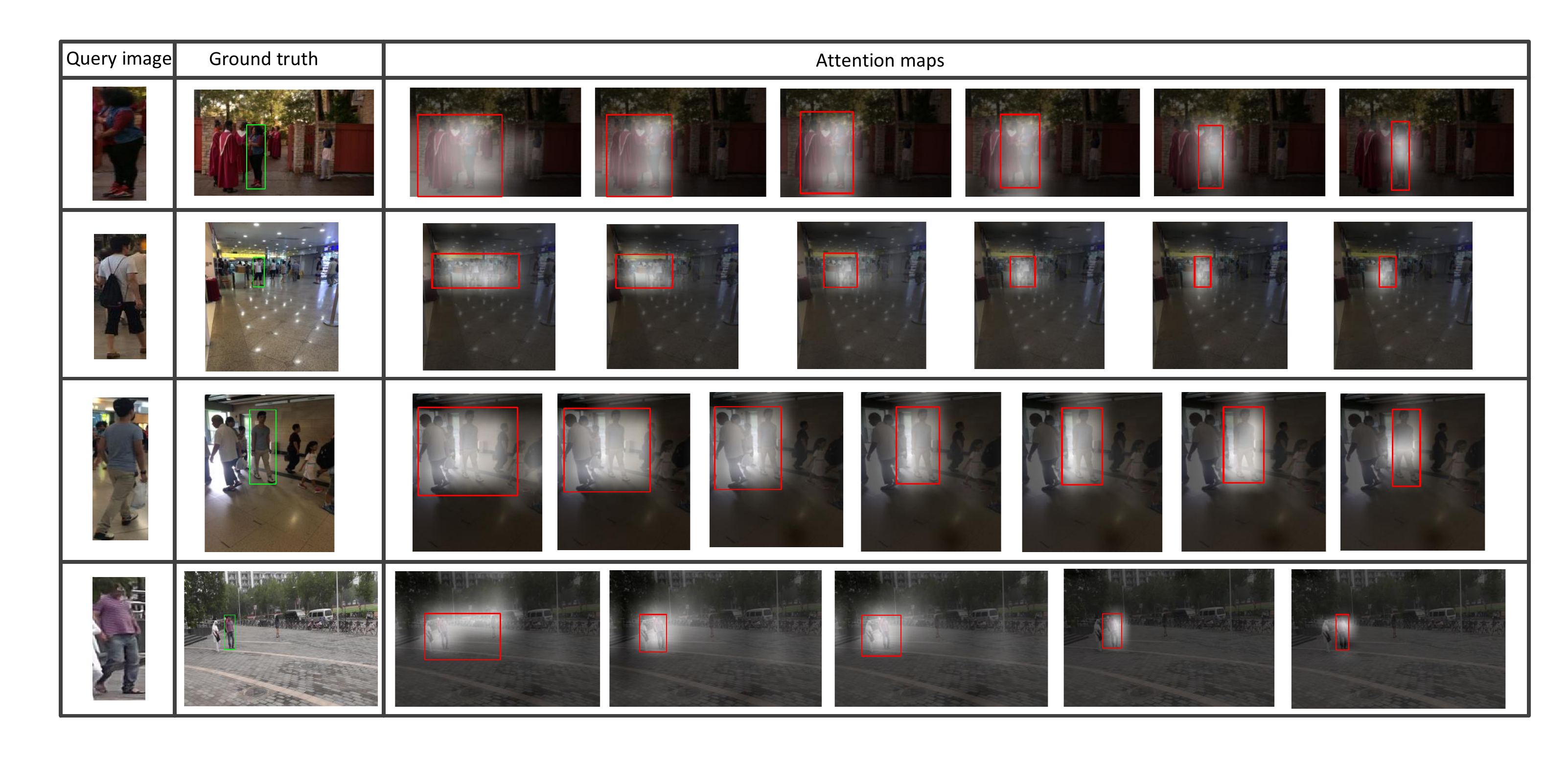}
	\vspace{-2mm}	
	\caption{Attention maps learned by our NPSM model for different testing person samples in CUHK-SYSU and PRW dataset. The first three rows are from CUHK-SYSU, while the bottom row is from PRW. Green boxes represent the ground truth boxes while red boxes are the region bounding boxes highlighted by our NPSM model.
	} 
	\vspace{-4.3mm}	
	\label{fig:attmap}
\end{figure*}

\subsubsection{Results on CUHK-SYSU}
We report the person search performance on CUHK-SYSU with 100 gallery size setting in Table \ref{tab:CUHK-SYSU}, where ``CNN'' represents the detector part (Faster-RCNN~\cite{ren2015faster} with ResNet-50) and ``IDNet" denotes the re-identification part in the framework of OIM\cite{xiao2016end}. Compared with CNN+IDNet, the OIM achieves performance improvement by introducing joint optimization of the detection and identification parts, but still follows the isolated ``detection+re-identification'' two-stage strategy in the person search process. Comparatively, our  proposed NPSM is a detection-free method and solves localization and re-identification of the query person simultaneously by introducing the query-aware region shrinkage mechanism which can include more contextual information beneficial for search accuracy. It can be verified by results shown in Table \ref{tab:CUHK-SYSU}. NPSM beats all compared methods consistently for both the mAP and top-1 metrics.

\vspace{-1.5mm}       
\begin{table}[h]
	\centering
	\small	
	\caption{Comparison of NPSM's performance on CUHK-SYSU with 100 gallery size setting with the state-of-the-arts.}
	\label{tab:CUHK-SYSU}%
	\begin{tabular}{p{4.2cm}|p{1cm}p{1.3cm}}
		\toprule
		\textbf{Method} & \textbf{mAP(\%)} & \textbf{top-1(\%)} \\
		\midrule
		ACF\cite{dollar2014fast}+DSIFT\cite{Zhao2013Unsupervised}+Euclidean  & 21.7 & 25.9 \\
		ACF+DSIFT+KISSME\cite{K2012large}  &32.3  &38.1  \\
		ACF+BoW\cite{Zheng2015Scalable}+Cosine & 42.4 & 48.4 \\
		ACF+LOMO+XQDA~\cite{Liao2015Person}&55.5&63.1\\
		ACF+IDNet\cite{xiao2016end} & 56.5 & 63.0 \\
		\midrule
		CCF\cite{Yang2015Convolutional}+DSIFT+Euclidean  & 11.3& 11.7 \\
		CCF+DSIFT+KISSME  &13.4  &13.9  \\
		CCF+BoW+Cosine & 26.9 & 29.3 \\
		CCF+LOMO+XQDA&41.2&46.4\\
		CCF+IDNet & 50.9& 57.1 \\
		\midrule
		CNN\cite{ren2015faster}+DSIFT+Euclidean  & 34.5& 39.4 \\
		CNN+DSIFT+KISSME  &47.8  &53.6  \\
		CNN+BoW+Cosine & 56.9 & 62.3 \\
		CNN+LOMO+XQDA&68.9&74.1\\
		CNN+IDNet & 68.6 & 74.8 \\
		OIM\cite{xiao2016end}(Baseline) &  75.5     & 78.7 \\
		\midrule
		\textbf{NPSM} & \textbf{77.9} & \textbf{81.2} \\
		\bottomrule
	\end{tabular}%
	\vspace{-3mm}
\end{table}%

Moreover, Figure~\ref{fig:gallerysize} shows the mAP of the compared methods with different gallery sizes, including [50, 100, 500, 1000, 2000, 4000]. One can see that the mAP drops gradually as the gallery size increases, but our method can still outperform all other methods under different gallery size settings. In particular, NPSM improves average performance per gallery size over OIM\cite{xiao2016end} by around 2\%. 

\vspace{-3.5mm}

\subsubsection{Results on PRW}
\vspace{-1mm}
On PRW dataset, we  conduct  experiments to compare NPSM with some state-of-the-art methods combining different detectors (respective R-CNN~\cite{girshick2016region} detectors of DPM~\cite{felzenszwalb2010object}, CCF\cite{Yang2015Convolutional},ACF~\cite{dollar2014fast}, LDCF~\cite{Nam2014Local}) and recognizers (LOMO, XQDA~\cite{Liao2015Person}, $\text{IDE}_{det}$, CWS~\cite{zheng2016person}). Among them,  AlexNet~\cite{krizhevsky2012imagenet} is exploited as the base network for the R-CNN detector. 
Although VGGNet~\cite{simonyan2014very} and  ResNet~\cite{he2016deep} have more parameters and are deeper than AlexNet, according to \cite{zheng2016person}, AlexNet can achieve better performance than the other two for DPM and ACF incorporating different recognizers. The results are shown in Table~\ref{tab:PRW}. Because the OIM method is the baseline of our NPSM, we implement the source code provided in OIM\cite{xiao2016end} to obtain the baseline result on the PRW dataset. Compared with the result, our NPSM  outperforms it by 2.9\% and 3.2\% for mAP and top-1 accuracy separately. Besides, compared with all other state-of-the-arts considering five bounding boxes per gallery image, our method  achieves better performance by only keeping one bounding box for testing per gallery image.

\begin{table}
	\centering
	\small	
	\caption{Comparison of NPSM's performance on PRW with state-of-the-arts.}
	\label{tab:PRW}%
	\begin{tabular}{l|cc}
		\toprule
		\textbf{Method} & \textbf{mAP(\%)} & \textbf{top-1(\%)} \\
		\hline
		$\text{DPM-Alex+LOMO+XQDA}$\cite{Liao2015Person}  & 13.0 & 34.1 \\
		$\text{DPM-Alex+IDE}_{det}$\cite{zheng2016person}  & 20.3 & 47.4 \\
		$\text{DPM-Alex+IDE}_{det}\text{+CWS}$\cite{zheng2016person}  & 20.5 & 48.3 \\    
		\midrule
		$\text{ACF-Alex+LOMO+XQDA}$  & 10.3 & 30.6 \\
		$\text{ACF-Alex+IDE}_{det}$  & 17.5 & 43.6 \\
		$\text{ACF-Alex+IDE}_{det}\text{+CWS}$  & 17.8 & 45.2 \\
		\midrule
		$\text{LDCF+LOMO+XQDA}$  & 11.0 & 31.1 \\
		$\text{LDCF+IDE}_{det}$  & 18.3 & 44.6 \\
		$\text{LDCF+IDE}_{det}\text{+CWS}$  & 18.3 & 45.5 \\
		OIM(Baseline) &21.3&49.9\\
		\midrule
		\textbf{NPSM} & \textbf{24.2} & \textbf{53.1} \\
		\bottomrule
	\end{tabular}%
	\vspace{-3mm}
\end{table}%

In Figure~\ref*{fig:attmap}, we visualize some attention maps
produced by our NPSM for testing samples from CUHK-SYSU and PRW datasets, which are all ranked top 1 in search results. The first three rows are from CUHK-SYSU, while the bottom row is from PRW. We observe that our NPSM can effectively shrink the search region to the correct person region guided by primitive memory of the query person.

\vspace{-1.5mm}
\section{Conclusions}
In this work, we introduced a novel neural person search machine solving person search  through recursively localizing effective regions, with guidance from the memory of the query person. Extensive experiments
on two public benchmarks demonstrated its superiority over state-of-the-arts in most
cases and the benefit to recognition accuracy in person search.

\section*{Acknowledgment}
This work was supported in part by the National Natural Science Foundation of China under Grant 61371155, Grant 61174170, and Grant 61632007. The work of Jiashi Feng was partially supported by NUS startup R-263-000-C08-133, MOE Tier-I R-263-000-C21-112 and IDS R-263-000-C67-646.

{\footnotesize
	\bibliographystyle{ieee}
	\bibliography{p_search}

\begin{thebibliography}{10}\itemsep=-1pt

\bibitem{ahmed2015improved}
E.~Ahmed, M.~Jones, and T.~K. Marks.
\newblock An improved deep learning architecture for person re-identification.
\newblock In {\em {IEEE CVPR}}, pages 3908--3916, 2015.

\bibitem{anderson1990cognitive}
J.~R. Anderson.
\newblock {\em Cognitive psychology and its implications}.
\newblock WH Freeman/Times Books/Henry Holt \& Co, 1990.

\bibitem{cai2015learning}
Z.~Cai, M.~Saberian, and N.~Vasconcelos.
\newblock Learning complexity-aware cascades for deep pedestrian detection.
\newblock In {\em {IEEE CVPR}}, pages 3361--3369, 2015.

\bibitem{dollar2014fast}
P.~Doll{\'a}r, R.~Appel, S.~Belongie, and P.~Perona.
\newblock Fast feature pyramids for object detection.
\newblock {\em {IEEE TPAMI}}, 36(8):1532--1545, 2014.

\bibitem{farenzena2010person}
M.~Farenzena, L.~Bazzani, A.~Perina, V.~Murino, and M.~Cristani.
\newblock Person re-identification by symmetry-driven accumulation of local
  features.
\newblock In {\em {IEEE CVPR}}, pages 2360--2367. IEEE, 2010.

\bibitem{felzenszwalb2010object}
P.~F. Felzenszwalb, R.~B. Girshick, D.~McAllester, and D.~Ramanan.
\newblock Object detection with discriminatively trained part-based models.
\newblock {\em {IEEE TPAMI}}, 32(9):1627--1645, 2010.

\bibitem{girshick2016region}
R.~Girshick, J.~Donahue, T.~Darrell, and J.~Malik.
\newblock Region-based convolutional networks for accurate object detection and
  segmentation.
\newblock {\em {IEEE TPAMI}}, 38(1):142--158, 2016.

\bibitem{gray2008viewpoint}
D.~Gray and H.~Tao.
\newblock Viewpoint invariant pedestrian recognition with an ensemble of
  localized features.
\newblock In {\em {ECCV}}, pages 262--275. Springer, 2008.

\bibitem{he2016deep}
K.~He, X.~Zhang, S.~Ren, and J.~Sun.
\newblock Deep residual learning for image recognition.
\newblock In {\em Proceedings of the IEEE Conference on Computer Vision and
  Pattern Recognition}, pages 770--778, 2016.

\bibitem{hochreiter1997long}
S.~Hochreiter and J.~Schmidhuber.
\newblock Long short-term memory.
\newblock {\em Neural computation}, 9(8):1735--1780, 1997.

\bibitem{jia2014caffe}
Y.~Jia, E.~Shelhamer, J.~Donahue, S.~Karayev, J.~Long, R.~Girshick,
  S.~Guadarrama, and T.~Darrell.
\newblock Caffe: Convolutional architecture for fast feature embedding.
\newblock {\em arXiv preprint arXiv:1408.5093}, 2014.

\bibitem{K2012large}
M.~K{\"o}stinger, M.~Hirzer, P.~Wohlhart, P.~M. Roth, and H.~Bischof.
\newblock Large scale metric learning from equivalence constraints.
\newblock In {\em {IEEE CVPR}}, pages 2288--2295, 2012.

\bibitem{krizhevsky2012imagenet}
A.~Krizhevsky, I.~Sutskever, and G.~E. Hinton.
\newblock Imagenet classification with deep convolutional neural networks.
\newblock In {\em {NIPS}}, pages 1097--1105, 2012.

\bibitem{li2014deepreid}
W.~Li, R.~Zhao, T.~Xiao, and X.~Wang.
\newblock Deepreid: Deep filter pairing neural network for person
  re-identification.
\newblock In {\em {IEEE CVPR}}, pages 152--159, 2014.

\bibitem{li2015multi}
X.~Li, W.-S. Zheng, X.~Wang, T.~Xiang, and S.~Gong.
\newblock Multi-scale learning for low-resolution person re-identification.
\newblock In {\em {IEEE ICCV}}, pages 3765--3773, 2015.

\bibitem{li2016videolstm}
Z.~Li, E.~Gavves, M.~Jain, and C.~G. Snoek.
\newblock Videolstm convolves, attends and flows for action recognition.
\newblock {\em arXiv preprint arXiv:1607.01794}, 2016.

\bibitem{Liao2015Person}
S.~Liao, Y.~Hu, X.~Zhu, and S.~Z. Li.
\newblock Person re-identification by local maximal occurrence representation
  and metric learning.
\newblock In {\em {IEEE CVPR}}, pages 2197--2206, 2015.

\bibitem{liu2016end}
H.~Liu, J.~Feng, M.~Qi, J.~Jiang, and S.~Yan.
\newblock End-to-end comparative attention networks for person
  re-identification.
\newblock {\em {IEEE TIP}}, 26(7):3492--3506, 2017.

\bibitem{liu2017video}
H.~Liu, Z.~Jie, K.~Jayashree, M.~Qi, J.~Jiang, S.~Yan, and J.~Feng.
\newblock Video-based person re-identification with accumulative motion
  context.
\newblock {\em arXiv preprint arXiv:1701.00193}, 2017.

\bibitem{liu2015kernelized}
H.~Liu, M.~Qi, and J.~Jiang.
\newblock Kernelized relaxed margin components analysis for person
  re-identification.
\newblock {\em IEEE Signal Processing Letters}, 22(7):910--914, 2015.

\bibitem{Nam2014Local}
W.~Nam, P.~Dollar, and J.~H. Han.
\newblock Local decorrelation for improved pedestrian detection.
\newblock {\em {NIPS}}, 1:424--432, 2014.

\bibitem{nino2016scalable}
J.~Ni{\~n}o-Casta{\~n}eda, A.~Fr{\'\i}as-Vel{\'a}zquez, N.~B. Bo,
  M.~Slembrouck, J.~Guan, G.~Debard, B.~Vanrumste, T.~Tuytelaars, and
  W.~Philips.
\newblock Scalable semi-automatic annotation for multi-camera person tracking.
\newblock {\em {IEEE TIP}}, 25(5):2259--2274, 2016.

\bibitem{Olshausen1993A}
B.~A. Olshausen, C.~H. Anderson, and D.~C. Van~Essen.
\newblock A neurobiological model of visual attention and invariant pattern
  recognition based on dynamic routing of information.
\newblock {\em Journal of Neuroscience the Official Journal of the Society for
  Neuroscience}, 13(11):4700--19, 1993.

\bibitem{ren2015faster}
S.~Ren, K.~He, R.~Girshick, and J.~Sun.
\newblock Faster r-cnn: Towards real-time object detection with region proposal
  networks.
\newblock In {\em {NIPS}}, pages 91--99, 2015.

\bibitem{sharma2015action}
S.~Sharma, R.~Kiros, and R.~Salakhutdinov.
\newblock Action recognition using visual attention.
\newblock {\em arXiv preprint arXiv:1511.04119}, 2015.

\bibitem{simonyan2014very}
K.~Simonyan and A.~Zisserman.
\newblock Very deep convolutional networks for large-scale image recognition.
\newblock {\em arXiv preprint arXiv:1409.1556}, 2014.

\bibitem{tao2016person}
D.~Tao, Y.~Guo, M.~Song, Y.~Li, Z.~Yu, and Y.~Y. Tang.
\newblock Person re-identification by dual-regularized kiss metric learning.
\newblock {\em {IEEE TIP}}, 25(6):2726--2738, 2016.

\bibitem{2016arXiv160502688short}
{Theano Development Team}.
\newblock {Theano: A {Python} framework for fast computation of mathematical
  expressions}.
\newblock {\em arXiv e-prints}, abs/1605.02688, May 2016.

\bibitem{tian2015deep}
Y.~Tian, P.~Luo, X.~Wang, and X.~Tang.
\newblock Deep learning strong parts for pedestrian detection.
\newblock In {\em {IEEE ICCV}}, pages 1904--1912, 2015.

\bibitem{tieleman2012lecture}
T.~Tieleman and G.~Hinton.
\newblock Lecture 6.5-rmsprop: Divide the gradient by a running average of its
  recent magnitude.
\newblock {\em COURSERA: Neural Networks for Machine Learning}, 4(2), 2012.

\bibitem{wu2016personnet}
L.~Wu, C.~Shen, and A.~v.~d. Hengel.
\newblock Personnet: Person re-identification with deep convolutional neural
  networks.
\newblock {\em arXiv preprint arXiv:1601.07255}, 2016.

\bibitem{xiao2016learning}
T.~Xiao, H.~Li, W.~Ouyang, and X.~Wang.
\newblock Learning deep feature representations with domain guided dropout for
  person re-identification.
\newblock {\em arXiv preprint arXiv:1604.07528}, 2016.

\bibitem{xiao2016end}
T.~Xiao, S.~Li, B.~Wang, L.~Lin, and X.~Wang.
\newblock Joint detection and identification feature learning for person
  search.
\newblock {\em arXiv:1604.01850}, 2017.

\bibitem{xingjian2015convolutional}
S.~Xingjian, Z.~Chen, H.~Wang, D.-Y. Yeung, W.-k. Wong, and W.-c. Woo.
\newblock Convolutional lstm network: A machine learning approach for
  precipitation nowcasting.
\newblock In {\em {NIPS}}, pages 802--810, 2015.

\bibitem{xu2015show}
K.~Xu, J.~Ba, R.~Kiros, K.~Cho, A.~Courville, R.~Salakhudinov, R.~Zemel, and
  Y.~Bengio.
\newblock Show, attend and tell: Neural image caption generation with visual
  attention.
\newblock In {\em {IEEE ICCV}}, pages 2048--2057, 2015.

\bibitem{Yang2015Convolutional}
B.~Yang, J.~Yan, Z.~Lei, and S.~Z. Li.
\newblock Convolutional channel features.
\newblock In {\em {IEEE ICCV}}, pages 82--90, 2015.

\bibitem{Yang2016Large}
Y.~Yang, S.~Liao, Z.~Lei, and S.~Z. Li.
\newblock Large scale similarity learning using similar pairs for person
  verification.
\newblock In {\em AAAI}, 2016.

\bibitem{zhang2016learning}
L.~Zhang, T.~Xiang, and S.~Gong.
\newblock Learning a discriminative null space for person re-identification.
\newblock {\em arXiv preprint arXiv:1603.02139}, 2016.

\bibitem{Zhao2013Unsupervised}
R.~Zhao, W.~Ouyang, and X.~Wang.
\newblock Unsupervised salience learning for person re-identification.
\newblock In {\em {IEEE CVPR}}, pages 3586--3593, 2013.

\bibitem{Zheng2015Scalable}
L.~Zheng, L.~Shen, L.~Tian, S.~Wang, J.~Wang, and Q.~Tian.
\newblock Scalable person re-identification: A benchmark.
\newblock In {\em {IEEE ICCV}}, pages 1116--1124, 2015.

\bibitem{zheng2016person}
L.~Zheng, H.~Zhang, S.~Sun, M.~Chandraker, and Q.~Tian.
\newblock Person re-identification in the wild.
\newblock {\em arXiv preprint arXiv:1604.02531}, 2016.

\bibitem{zitnick2014edge}
C.~L. Zitnick and P.~Doll{\'a}r.
\newblock Edge boxes: Locating object proposals from edges.
\newblock In {\em {ECCV}}, pages 391--405. Springer, 2014.

\end{thebibliography}
}

\end{document}